\documentclass{article}
\usepackage{ijcai26}

\usepackage{times}
\usepackage{soul}
\usepackage{url}
\usepackage[hidelinks]{hyperref}
\usepackage[utf8]{inputenc}
\usepackage[small]{caption}
\usepackage{graphicx}
\usepackage{amsmath}
\usepackage{amsthm}
\usepackage{booktabs}
\usepackage{algorithm}
\usepackage{algorithmic}
\usepackage[switch]{lineno}
\usepackage{tablefootnote}
\usepackage{booktabs}
\usepackage{tabularx}
\usepackage{threeparttable}
\usepackage{hyperref}
\usepackage{xurl}
\usepackage{booktabs}      
\usepackage{tablefootnote} 
\usepackage{pifont}        
\usepackage[textsize=tiny]{todonotes}
\usepackage{amssymb}

\newcommand{\cmark}{\ding{51}} 
\newcommand{\xmark}{\ding{55}} 

\title{Graph Rewiring in GNNs to Mitigate Over-Squashing and Over-Smoothing: A Survey}

\author{
Hugo Attali$^1$
\and
Nathalie Pernelle$^1$\and
Davide Buscaldi$^{1}$\And
Fragkiskos D. Malliaros$^2$\\
\affiliations
$^1$Université Sorbonne Paris Nord, CNRS, LIPN\\
$^2$Université Paris-Saclay, CentraleSupélec, Inria\\
\emails
\{attali, pernelle, buscaldi\}@lipn.univ-paris13.fr, fragkiskos.malliaros@centralesupelec.fr
}
\begin{document}

\maketitle

\begin{abstract}

Graph Neural Networks are powerful models for learning from graph-structured data, yet their effectiveness is often limited by two critical challenges: over-squashing, where information from distant nodes is excessively compressed, and over-smoothing, where repeated propagation makes node representations indistinguishable. Both phenomena stem from the interaction between message passing and the input topology, ultimately degrading information flow and limiting the performance of GNNs. In this survey, we examine graph rewiring techniques, a class of methods designed to modify the graph topology to enhance information propagation in GNNs. We provide a comprehensive review of state-of-the-art rewiring approaches, delving into their theoretical underpinnings, practical implementations, and performance trade-offs.

\end{abstract}
\section{Introduction }

Graph Neural Networks (GNNs) \cite{gori2005new,scarselli2008graph,gilmer2017neural} have emerged as powerful models for analyzing structured data represented as graphs across various fields \cite{zhou2020graph}. However, they face significant challenges, especially in heterophilic graphs where nodes belonging to different classes are highly interconnected, and more generally in problems defined on graphs with pronounced long-range dependencies, where task-relevant information may lie several hops away from each node. In such scenarios, the ability to propagate long-range information becomes crucial, yet it is often hindered by phenomena such as over-squashing \cite{alon2020bottleneck,UNDERSTANDING_bottlenecks,arnaiz-rodriguez2026oversmoothing} and over-smoothing \cite{rusch2023survey}. Beyond architectural changes, a growing line of work intervenes directly on the propagation operator used by message passing. A first family of approaches relies on {edge reweighting, where existing connections are assigned adaptive strengths to better reflect latent affinities between nodes and to attenuate noisy edges \cite{zheng2022graph}. While this can improve local aggregation, such reweighting remains constrained by the original graph support and therefore offers limited leverage in heterophilic or long-range settings. To explicitly increase the accessible dependency range, complementary work has studied multi-hop or higher-order aggregation strategies \cite{gong2026survey}. These methods extend each node’s receptive field by replacing one-step neighborhoods with higher-order connectivity (e.g., powers of the adjacency matrix), thereby enabling interactions between distant nodes. From a graph perspective, they induce a topology by transforming the adjacency matrix and can thus be interpreted as an implicit form of rewiring. However, the benefits of 
multi-hop strategies come with a trade-off: aggregating over large hop neighborhoods can intensify information compression and accentuate over-squashing \cite{alon2020bottleneck,UNDERSTANDING_bottlenecks}, while the graph densification may accelerate over-smoothing and hinder scalability on large graphs \cite{BORF,giraldo2023trade}. These limitations motivate graph rewiring that modifies the graph topology to promote long-range information flow while attenuating structural patterns that induce excessive mixing or compression, and thereby exacerbate over-smoothing and over-squashing. Complementing the over-squashing survey of \cite{akansha2025over}, this work provides a method-centric survey of rewiring techniques, covering both preprocessing and dynamic rewiring methods to mitigate over-squashing and over-smoothing.

\paragraph{Objectives and outline of the survey.} Our goal is to provide a structured overview of rewiring approaches, with an emphasis on their underlying principles,  design choices, and trade-offs they introduce. We highlight why such techniques are particularly relevant in challenging regimes, including heterophilic graphs and tasks exhibiting long-range dependencies,  where informative signals may lie beyond the immediate neighborhood.
Finally, we outline open problems and future directions, positioning graph rewiring as a principled way to intervene on graph structure in order to better understand and improve graph learning. We argue that progress will benefit from clearer problem formulations, more explicit assumptions, and evaluation protocols that make claims robust, comparable, and interpretable across settings.

\section{Preliminary Concepts}
\subsection{Notation}
We begin by introducing the notation used throughout this paper. A graph is represented as a tuple \({G}=(\mathcal{V}, \mathcal{E})\), where \(\mathcal{V}\) denotes the set of nodes and \(\mathcal{E}\) the set of edges. The number of nodes is denoted by \(N=|\mathcal{V}|\), and an edge connecting node \(i\) to node \(j\) is represented by \(e_{ij} \in \mathcal{E}\). In this work, we focus on undirected graphs, meaning that if \(e_{ij} \in \mathcal{E}\), then \(e_{ji} \in \mathcal{E}\). We define the \(N \times N\) adjacency matrix \(\mathbf{A}\) such that \(\mathbf{A}_{i,j}=1\) if \((i,j) \in \mathcal{E}\), and 0 otherwise. Additionally, \(\mathbf{D}\) denotes the diagonal matrix where \(\mathbf{D}_{i,i}=d_i\), the degree of node \(i\). The maximum and minimum degrees are denoted by \(d_{\max}\) and \(d_{\min}\), respectively. We note by $ h_{i}^{(\ell)}$ the embedding of the node $i$ at the $\ell$ layer.


\subsection{Graph Neural Networks}
Graph Neural Networks (GNNs) are a class of models designed to leverage the structure of graph data by iteratively propagating and updating node features through their neighborhoods. This process is typically framed within the Message Passing Neural Networks (MPNNs) paradigm, where node representations are refined by exchanging information with neighboring nodes. Each iteration of the GNN involves two operations, aggregation and update to compute embeddings  $h_{i}^{\ell}$ at the layer $\ell$ based on message $m_{i}^{(\ell)}$ containing information on neighbors $\mathcal{N(i)}$, as follows: 
\begin{equation}
    \begin{split}
         & m_{i}^{(\ell)} = \texttt{AGGREGATE}^{(\ell)}\left(h_i^{(\ell-1)},\left\{ h_{j}^{(\ell-1)} \hspace{1mm} j \in \mathcal{N}(i)\right\}\right), \\
         & h_{i}^{(\ell)} = \texttt{UPDATE}^{(\ell)}\left(h_i^{(\ell-1)}, m_{i}^{(\ell)}\right). 
    \end{split}
\label{aggreg}
\end{equation}

\subsection{Heterophily and Long-range Dependencies}

We call a graph heterophilic when adjacent nodes are not likely to share the same label \cite{Homophily}. In this regime, the 1-hop neighborhood could be a misleading predictor of a node's label: local aggregation mixes signals coming from different classes, so useful evidence may emerge only after aggregating information over longer paths, or from more distant regions that are predictive of the target label.

Long-range dependencies refer to tasks where the target at a node (or graph) depends on information located at a large graph distance. In this context, since MPNNs propagate information locally (one hop per layer), capturing signals located at large graph distances typically requires increasing depth. Greater depth, however, tends to amplify two well-known limitations of message passing: (i) over-squashing, i.e., the aggregation of many distinct long-range signals into fixed-size node representations, and (ii) over-smoothing, i.e., the progressive loss of discriminative power caused by repeated neighborhood mixing. Consequently, heterophilic settings and long-range dependency tasks are particularly sensitive to these effects.

\subsection{Over-squashing}


Recent studies have highlighted that message passing neural networks struggle with tasks that require long-range dependencies \cite{di2023over}. A useful way to formalize this limitation is through long-range influence, quantified by a Jacobian-based sensitivity: for two nodes $i$ and $j$, let
\[
J^{(\ell)}_{i \leftarrow j} \;=\;\Bigl\|\tfrac{\partial h_i^{(\ell)}}{\partial h_j^{(0)}}\Bigr\|
\]
denote the Jacobian sensitivity of $h_i^{(\ell)}$ to perturbations of $h_j^{(0)}$.
Over-squashing denotes the regime where this effect decays rapidly with graph distance, so that signals from an exponentially growing $k$-hop neighborhood are compressed into fixed-size representations \cite{di2023over}.
Over-squashing is amplified by the graph structure. If two far-apart regions are linked by only a few paths, most long-range messages must pass through the same small set of paths, which increases compression and makes distant information harder to preserve. These structural bottlenecks thus limit long-range communication in message passing \cite{UNDERSTANDING_bottlenecks}.

\subsection{Over-smoothing}
Over-smoothing is a complementary depth-related limitation of message passing \cite{rusch2023survey}. Since each layer repeatedly mixes information across neighborhoods, deep propagation behaves like a diffusion process that progressively reduces variation across node representations. One standard way to formalize over-smoothing is via the asymptotic collapse of pairwise differences:
\[
\max_{i,j}\bigl\|h_i^{(\ell)}-h_j^{(\ell)}\bigr\|\;\longrightarrow\;0
\qquad(\ell\to\infty),
\]
meaning that node embeddings become increasingly indistinguishable and thus less informative for discrimination \cite{Over-Smoothing_gnn}. In practice, this homogenization can be intensified when the graph becomes denser, as stronger mixing accelerates the loss of feature diversity.





\subsection{Rewiring}
Since these limitations are largely exacerbated by the graph topology, a natural approach is to modify the graph on which message passing is performed to mitigate them.

\paragraph{\textbf{Definition.}}
Let $G=(V,E)$ be a simple undirected graph. A \emph{rewiring} is any mapping
\[
\mathcal{R}:\ \mathcal{G}\to\mathcal{G},\qquad
G\ \mapsto\ G^{+}=(V^{+},E^{+}),
\]
where, possibly, $V\subseteq V^{+}$ (augmentation with virtual nodes) and where $E^{+}$ is typically obtained from $E$ by adding and/or removing edges .
We call a rewiring static if $\mathcal{R}$ is applied once prior to training. It is dynamic if the rewired graph depends on learnable parameters and/or evolves over training, i.e., $G^{+}=\mathcal{R}_{\theta}(G)$ where $\theta$ denotes the learnable parameters of the GNN used during training.

The objective is to obtain a modified graph \( G^{+} \) whose topology is more favorable to information propagation under an MPNN than that of the original graph \( G \). 
\noindent Unlike traditional graph learning methods that typically adjust the \texttt{UPDATE} and/or \texttt{AGGREGATE} operations while maintaining fixed neighborhoods, 
rewiring methods approach the problem differently. They treat graph modification as allowing alterations to the neighborhood structure while keeping the message passing operations intact. Figure \ref{fig:overload} presents an example of heuristic graph rewiring.


\begin{figure}[t]
    \centering
    \includegraphics[scale=0.45]{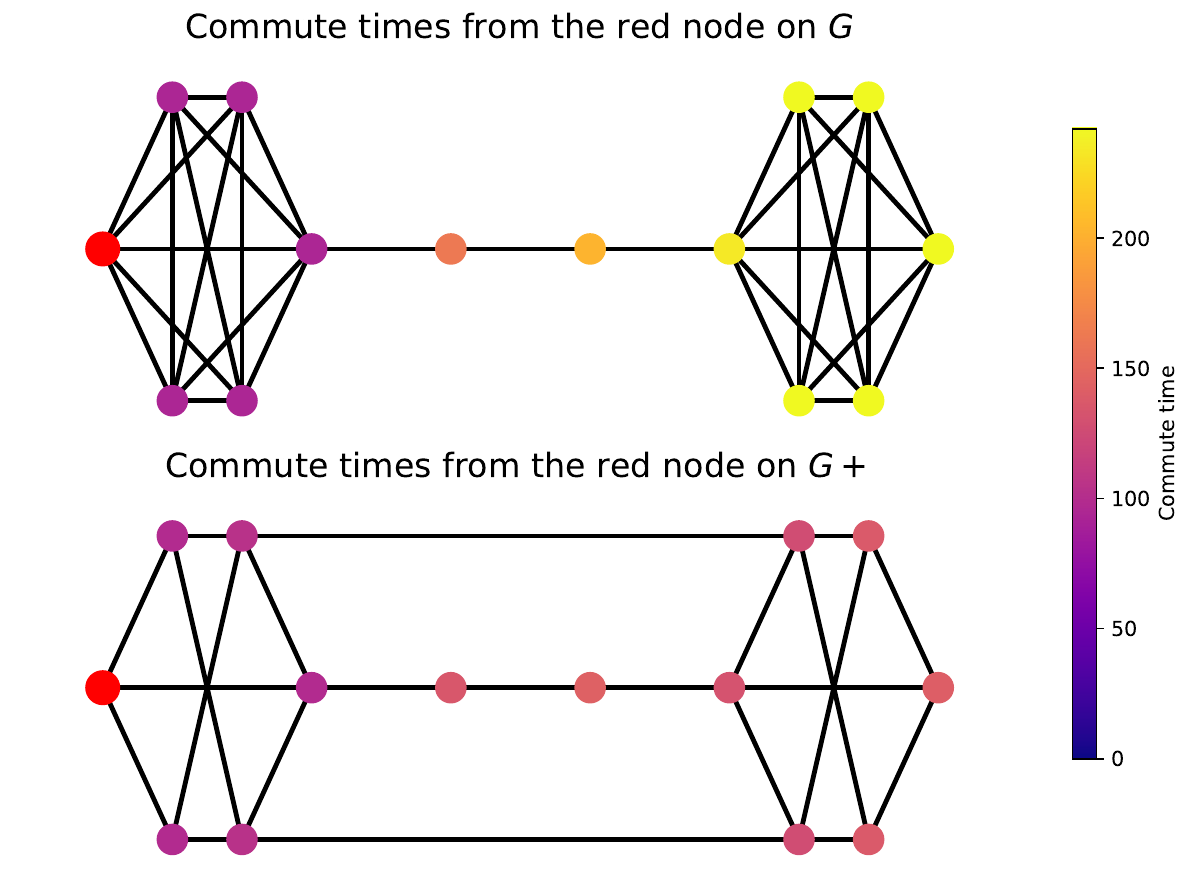}
    \caption{Example of a graph \(G\) and its rewired instance \(G^{+}\), with node colours encoding commute times (see Section \ref{RW})  from the highlighted red node. In \(G^{+}\), intra-clique density is reduced while connectivity across the bottleneck is strengthened, which shortens and balances commute times. This yields a topology more favourable to information propagation, thereby mitigating over-smoothing and over-squashing.
}
    \label{fig:overload}
    
\end{figure}

\section{Topological and Connectivity Measures}
\label{topo}
Various topological measures have been utilized and proposed as a basis for constructing \( G^{+} \).

\subsection{Discrete Curvature on Graphs}
\label{curv}

A prominent concept in graph rewiring is discrete curvature. Discrete curvature draws the analogy between dispersion geodesics on a manifold and an edge on a graph. 
Intuitively, edge curvature reflects the local structure around the neighborhoods of two connected nodes. Edge curvature was first linked to bottleneck structures by \cite{UNDERSTANDING_bottlenecks}, demonstrating that highly negatively curved edges characterise bottleneck structures.
Various discrete curvature measures have been studied for their potential to improve graph rewiring techniques.

\paragraph{\textbf{Ollivier Curvature.}}
\label{sec:ollivier}
Ollivier curvature is a discrete analogue of Ricci curvature based on optimal transport, measuring how similar the probability distributions supported on the local neighborhoods of two nodes are \cite{Olivier_curvature}. Let $\mu_i$ be the probability measure associated with node $i$, defined via a lazy random walk with parameter $\alpha$:
\begin{equation} 
\mu_i(j) =
\begin{cases}
\alpha & \text{if } j = i, \\
\dfrac{1-\alpha}{d_i} & \text{if } j \in \mathcal{N}(i), \\
0 & \text{otherwise},
\end{cases}
\end{equation}
where $d_i$ is the degree of node $i$ and $\mathcal{N}(i)$ is its set of neighbors. 
Let $W_1(\mu_i,\mu_j)$ denote the $1$-Wasserstein distance between $\mu_i$ and $\mu_j$, that is, the minimal transport cost between these two probability measures when the ground cost is given by the graph shortest-path distance. The Ollivier curvature $c^{\text{O}}_{ij}$ of an edge $e_{ij}$ is then defined as
\begin{equation}
\mathbf{C}^\text{O}_{i,j} = 1 - \frac{W_1(\mu_i, \mu_j)}{\text{dist}(i,j)},
\end{equation}
where $\text{dist}(i,j)$ denotes the shortest-path distance between nodes $i$ and $j$. Unlike other curvature measures, Ollivier curvature is  bounded, with $c^\text{O}_{ij} \in [-2, 1]$ \cite{BORF}, which can make it more interpretable.

\paragraph{\textbf{Augmented Forman Curvature.}}
Proposed by \cite{samal2018comparative}, this measure extends the original definition of Forman Curvature \cite{Forman} to incorporate the presence of triangles in a graph. For an undirected graph, the curvature of edge $e_{ij}$ is given by:
\begin{equation}
\mathbf{C}^\text{AF}_{i,j} = 4 - d_i - d_j + 3m,
\end{equation}
\noindent where $m$ is the number of triangles containing $e_{ij}$, and $d_i$ and $d_j$ are the degrees of nodes $i$ and $j$, respectively.

\paragraph{\textbf{Balanced Forman Curvature.}}
\cite{UNDERSTANDING_bottlenecks} introduced a more expressive combinatorial curvature measure, known as balanced Forman curvature, which considers not only triangles but also cycles of different lengths:
{\small
\begin{equation}
\begin{split}
\mathbf{C}^\text{BF}_{i,j} = \frac{2}{d_i} + \frac{2}{d_j} - 2 + 2 \frac{m}{\max \{d_i, d_j\}} + \\
\frac{m}{\min \{d_i, d_j\}} + \frac{(\Gamma_{\max})^{-1}}{\max \{d_i, d_j\}} (\gamma_{i} + \gamma_{j}),
\end{split}
\end{equation}}
\noindent where $\Gamma_{\max}(i,j)$ is the maximum number of 4-cycles based at edge $e_{ij}$, and $\gamma_i$ counts the number of 4-cycles at $e_{ij}$ without diagonals.

\paragraph{\textbf{Jost-Liu Curvature.}}
The Jost-Liu curvature further refines edge-based curvature measures by incorporating the local clustering around an edge, through the number of
triangles (3-cycles) containing $e_{ij}$ \cite{jost2014ollivier}:
{\small
\begin{equation}
\begin{aligned}
\mathbf{C}^\text{JL}_{i,j} &= -\left(1 - \frac{1}{d_i} - \frac{1}{d_j} - \frac{m}{d_i \wedge d_j}\right)_{+} \\
& \quad -\left(1 - \frac{1}{d_i} - \frac{1}{d_j} - \frac{m}{d_i \vee d_j}\right)_{+} + \frac{m}{d_i \vee d_j},
\end{aligned}
\end{equation}}
where $d_i$ and $d_j$ are the degrees of the nodes.  A more detailed discussion on the role of different edge curvature 
measures for capturing over-squashing information is provided 
in \cite{tori2024effectiveness,chen2026rethinking}.


\subsection{Effective Resistance} Effective resistance offers a more global perspective of how well two nodes communicate by considering all possible paths between them, rather than focusing solely on their local neighborhoods. It is defined as the inverse of the sum of the inverses of the lengths of all disjoint paths connecting two nodes, placing more weight on shorter paths \cite{chandra1989electrical}. The higher the effective resistance of an edge, the more it acts as a bottleneck in the graph. For vertices $i$ and $j$ connected by several disjoint paths, the effective resistance is given by: 
\begin{equation} \mathbf{R}_{i,j} = \left( \sum_{\text{$ij$-paths } p} \text{length}(p)^{-1} \right)^{-1}. \end{equation} 
A lower  $\mathbf{R}_{i,j}$ indicates easier communication and information flow between nodes $i$ and $j$, making it a valuable metric for identifying key structural bottlenecks in the graph.
Compared to curvature-based measures, effective resistance provides a global view of graph connectivity.

\subsection{Random Walk-based Measures}
\label{RW}
Walk-based quantities provide natural proxies for long-range connectivity, such a hitting times or return probabilities, and can be related to over-squashing by quantifying the ease with which information can propagate between distant regions of the graph. Commute time is particularly informative: it measures the expected number of steps for a random walk to travel from $i$ to $j$ and back, so lower commute times indicate easier bidirectional communication between distant nodes (see Fig.~\ref{fig:overload}). In the context of MPNNs, the number of random walks of a given length between two nodes controls how easily information can flow between them, especially at long distances. Theorem~4.1 in \cite{di2023over} shows that this directly impacts over-squashing: distant nodes with few connecting walks exchange information only weakly. For nodes $i$ and $j$ at distance $r$ in a graph $G$, let $\gamma_r(i, j)$ denote the number of random walks of length $r$ between $i$ and $j$. The theorem states that the Jacobian sensitivity between nodes decays with distance at a rate governed both by model parameters (Lipschitz constant $c_\sigma$, weights $w$, hidden dimension $p$) and graph properties (minimum degree $d_{\min}$, and walk counts $\gamma_r(i, j)$):
\begin{equation}
\frac{\partial h_i}{\partial h_j} \;\leq\; C_k \,\gamma_{r+k}(i, j)\,\left( \frac{2 c_\sigma w p}{d_{\min}} \right)^{r}.
\end{equation}
For large $r$, if $\gamma_{r+k}(i, j)$ is small, the Jacobian decays exponentially, which increases the risk of over-squashing. Equivalently, a large commute time between $i$ and $j$ on the graph, i.e., the expected number of steps needed to go from $i$ to $j$ and back, indicates poor connectivity and, consequently, stronger attenuation of messages between these nodes.

Building on this perspective, \cite{barbero2024localityaware} proposes a local connectivity measure
\begin{equation}
\mathbf{\Omega}_k(i,j) = \bigl(\tilde{\mathbf{A}}^k\bigr)_{i,j}, \quad \tilde{\mathbf{A}} = \mathbf{A} + \mathbf{I},
\end{equation}
which counts the number of walks from $i$ to $j$ of length $k$. For fixed $k$, a large $\mathbf{\Omega}_k(i,j)$ (or, relatedly, a small commute time) indicates strong connectivity and many routes for information exchange, whereas low $\mathbf{\Omega}_k(i,j)$ correlates with few available paths, larger commute times, and a higher susceptibility to over-squashing.

\subsection{Spectral Gap}

Another widely used measure is the spectral gap, which is closely related to the Cheeger constant \cite{chung1997spectral} defined as:
\begin{equation}
\mathrm{Ch}(G)=
\min_{\substack{\mathcal{S}\subset \mathcal V\\ 1\le |\mathcal{S}|\le |\mathcal V|/2}}
\frac{|\partial \mathcal{S}|}{|\mathcal{S}|}.
\end{equation}

where $\mathcal{S} \subset \mathcal{V}$ and $\partial \mathcal{S} = \{(i,j)\in \mathcal{E} : i\in \mathcal{S},\ j\in \mathcal{V}\setminus \mathcal{S}\}$ denotes the edge boundary of $\mathcal{S}$. Small values of $Ch(G)$ indicate the presence of a sparse cut, i.e., two large vertex groups connected by few edges, revealing a bottleneck; conversely, large values suggest that every balanced partition has many crossing edges. Computing $Ch(G)$ exactly is intractable in general, but it can be bounded using the second smallest eigenvalue $\lambda_2$ of the Laplacian matrix $\mathbf{L}$ via the discrete Cheeger inequality \cite{alon1984eigenvalues,cheeger2015lower}:
\begin{equation}
\frac{\lambda_2}{2} \le Ch(G) \le \sqrt{2\lambda_2}.
\end{equation}
Hence, a small $\lambda_2$ (small spectral gap) provides an efficient indicator of a near-disconnected structure: only a few edges may separate the graph into large components, which is precisely the type of bottleneck that can hinder information propagation in message passing GNNs.

\section{Rewiring Methods in GNNs}

Graph rewiring modifies the graph connectivity used for message passing to improve information propagation, typically with the goal of mitigating over-squashing and over-smoothing. Building on the bottleneck perspective of \cite{alon2020bottleneck}, later methods propose algorithmic changes to the graph, most commonly by adding, removing, or reweighting connections either locally to relieve specific structural bottlenecks or globally to reshape the overall topology. We categorize rewiring methods into three categories:

\begin{itemize}

\item \textbf{Type of structural modification}: the structural operator defining how $G$ is altered, whether through selective edge editing, global adjacency reconstruction, or augmentation via auxiliary nodes.

\item \textbf{Rewiring signal and information used}: the information source defining the rewiring rule (structure, features, or learned signals) and the scale at which it operates (local or global).

\item \textbf{Computational efficiency and hyperparameter sensitivity}: the computational overhead of the method and its robustness to rewiring hyperparameters.

\end{itemize}

Table \ref{Review} provides a summary of the primary rewiring methods that have been published to date. For each method, the table indicates whether it mitigates over-squashing (OSQ) and/or over-smoothing (OSM)  and specifies the underlying measures upon which the rewiring technique is based. Additionally, we indicate whether the graph rewiring method uses node features during rewiring (UF) and whether rewiring is applied during training (DT).

\subsection{Structural-fix Rewiring Strategies}
Let \(G=(\mathcal V,\mathcal E)\) be a graph and let
\(\mathbf{S_G}:\mathcal V\times\mathcal V\to \mathbb R^p\)
denote any topological or connectivity descriptor (e.g., curvature,
effective resistance, walk-based or spectral measures as in Section~\ref{topo})
computed from \(G\) alone.
A structural-fix rewiring method is a mapping
\[
\mathcal R:\mathcal G\to\mathcal G,\qquad
G\mapsto G^{+}=(\mathcal V^{+},\mathcal E^{+}),
\]
such that there exists a rule \(\Phi:\mathbb R^p\to\{0,1\}\) with
\[
(i,j)\in\mathcal E^{+}\ \Longleftrightarrow\ \Phi\big(\mathbf{S_G}(i,j)\big)=1,
\]
i.e., edges are added or removed solely as a function of topological descriptors, without using node features, labels, or model parameters.

\paragraph{\textbf{Curvature-based structural-fix methods.}} 
Let \(\mathbf{C}_{ij}\) denote the curvature of an edge \(e_{ij}\) as defined in Section~\ref{curv}.
Curvature-based structural-fix rewiring is defined as 
\[
\mathbf{s}_G:\mathcal{V}\times \mathcal{V}\to\mathbb{R},
\qquad
\mathbf{s}_G(i,j)=\mathbf{C}_{i,j}\quad\text{for }(i,j)\in\mathcal{E},
\]
together with a decision rule \(\Phi:\mathbb{R}\to\{0,1\}\) acting on curvature values
\cite{UNDERSTANDING_bottlenecks,fesser2023mitigating,giraldo2023trade,BORF,attali2025curvature}.
A typical instance fixes thresholds \(\tau_{\mathrm{neg}}<0<\tau_{\mathrm{pos}}\), to remove highly positively curved
edges ($\tau_{\mathrm{pos}}$), and then adds the minimum number of edges to alleviate highly negatively curved edges ($\tau_{\mathrm{neg}}$).
We introduce two sets: a deletion set \(\mathcal E_{\mathrm{del}}\subseteq \mathcal E\) and an addition
set \(\mathcal E_{\mathrm{add}}\subseteq \overline{\mathcal E}:=(\mathcal V\times\mathcal V)\setminus\mathcal E\).

\noindent Finally, we construct \(G^{+}=(\mathcal V,\mathcal E^{+})\) by jointly solving
\[
\begin{aligned}
(\mathcal E_{\mathrm{del}}^{\star},\mathcal E_{\mathrm{add}}^{\star})
\in \arg\min_{\mathcal E_{\mathrm{del}}\subseteq\mathcal E,\ \mathcal E_{\mathrm{add}}\subseteq\overline{\mathcal E}}
\ &|\mathcal E_{\mathrm{del}}|+|\mathcal E_{\mathrm{add}}|\\
\text{s.t.}\quad
&\mathcal E^{+}=(\mathcal E\setminus \mathcal E_{\mathrm{del}})\cup \mathcal E_{\mathrm{add}},\\
&\tau_{\mathrm{neg}}\le \mathbf C^{(\mathcal V,\mathcal E^{+})}_{ij}\le \tau_{\mathrm{pos}},\\
&\hspace{2.05em}\forall (i,j)\in \mathcal E^{+}.
\end{aligned}
\]

\noindent yielding \(\mathcal E^{+}:=(\mathcal E\setminus \mathcal E_{\mathrm{del}}^{\star})\cup \mathcal E_{\mathrm{add}}^{\star}\). This yields a rewired graph \(G^{+}\), where deletions target highly positive-curvature edges (mitigating over-smoothing) and additions create shortcuts across bottlenecks (mitigating over-squashing).
Note that exists also methods may add edges only (set $\tau_{\mathrm{pos}}=+\infty$) or delete edges only (set $\tau_{\mathrm{neg}}=-\infty$)\cite{attali2025curvature}.

\paragraph{\textbf{Resistance-based methods.}}

Curvature offers a local view of connectivity. 
Resistance-based methods instead rely on the effective resistance \(\mathbf{R}^{G}_{i,j}\), which aggregates all paths between \(i\) and \(j\) and yields a global connectivity descriptor. 
In the structural-fix framework, one sets
\[
s_G  = \mathbf{R}^{G}_{i,j}, \qquad (i,j)\in\mathcal{V}\times\mathcal{V}, 
\] 
and seeks a rewired graph \(G^{+} = (\mathcal{V},\mathcal{E}^{+})\) that minimizes a global resistance functional, typically
\[
G^{+} \in \arg\min_{\widetilde{G}} \sum_{i<j} \mathbf{R}^{\widetilde{G}}_{i,j}
\quad \text{subject to an edge budget on } \widetilde{\mathcal{E}},
\] 
as in \cite{resit}. In practice, the method (GTR) iteratively adds edges between node pairs with large $\mathbf{R}^{G}_{i,j}$, strengthening long-range connectivity by bringing them closer in effective-resistance distance.

\paragraph{\textbf{Walk  methods.}}
Because both curvature and effective resistance can be computationally demanding, walk-based descriptors have been proposed as cheaper proxies. 
For a fixed walk length \(k\), one may define :
\[
\mathbf{\Omega}_G(i,j) = \sum_{\ell=1}^{k} (\mathbf{A}^{\ell})_{ij},
\]
i.e., the number of walks of length at most \(k\) between \(i\) and \(j\), where \(\mathbf{A}\) is the adjacency matrix of \(G\). 
A structural-fix rule then promotes edges between pairs with few walks (low \(s_G (i,j)\)) and deprioritizes pairs with many alternative routes, consistent with the observation that nodes connected by many walks are less sensitive to over-squashing \cite{di2023over}. 
In practice, this principle motivates schemes based on adjacency powers, such as LASER’s \(k\)-power construction \cite{barbero2024localityaware}.

\paragraph{\textbf{Spectral gap  methods.}}
Spectral methods act at a global scale. 
Let \(\mathbf{L}\) denote a normalized Laplacian of \(G\), and let
$\gamma(G) = \lambda_2(L)$ be its spectral gap. 
Since a larger spectral gap is closely related to lower effective resistance and better expansion, spectral-gap based rewiring aims to construct a graph \(G^{+}\) with increased \(\gamma(G^{+})\). 
Formally, one considers
{\small
\[
G^{+}\in\arg\max_{\widetilde G}\ \gamma(\widetilde G)
\quad \text{subject to an edge-budget constraint on }\widetilde{\mathcal E}.
\]
}
\noindent and approximates this objective by iteratively adding edges that most improve the spectral gap (FOSR) \cite{FOSR}. 
Expander-based constructions \cite{deac2022expander,wilson2024cayley} are a canonical example of this principle:
expanders achieve large spectral gap, small diameter, and high Cheeger constant with relatively few edges, thereby reducing bottlenecks and strongly mitigating over-squashing.
More recent approaches such as spectral graph pruning \cite{jamadandi2024spectral} and spectrum preserving sparsification \cite{liang2025mitigating} refine this idea by assigning a spectral importance score to each edge and selecting a sparse subgraph that, respectively, improves the spectral gap or approximately preserves the Laplacian spectrum, while still providing favourable conditions for long-range propagation.

\paragraph{\textbf{Virtual node methods.}}
In addition to modifying the edge set, some structural-fix methods augment the node set with auxiliary nodes. 
A canonical example is the master node: one considers the augmented graph 
\(G^{+} = (\mathcal{V} \cup \{M\}, \mathcal{E}^{\prime})\), where 
\[
\mathcal{E}^{\prime} = \mathcal{E} \cup \{(M,i) : i \in \mathcal{V}\},
\]
so that every original node is connected to \(M\). 
This construction reduces the diameter of the original node set to \(2\), decreases effective resistance between distant nodes, and thereby mitigates over-squashing \cite{southern2025understanding}. 
More recent fractal-node constructions introduce a collection of auxiliary nodes 
\(\mathcal{F}\) together with attachment sets \(\{S_f \subseteq \mathcal{V} : f \in \mathcal{F}\}\), yielding
\[
G^{+} = (\mathcal{V} \cup \mathcal{F},\, \mathcal{E} \cup \{(f,i) : f \in \mathcal{F},\, i \in S_f\}),
\]
in order to create multi-scale shortcut structures that facilitate long-range communication without enforcing a single global hub topology \cite{choi2025graph}.

\subsection{Feature-aware Rewiring Strategies}

Beyond structural-fix rewiring, feature-aware methods use node attributes as the rewiring signal, which is particularly relevant when the observed topology is noisy, incomplete, or weakly aligned with the task.

Delaunay Rewiring (DR) \cite{attali2024delaunay}  discards $\mathcal{E}$ and reconstructs $G^{+}$ via Delaunay triangulation in a two-dimensional feature space, yielding a sparse planar graph. This triangulation avoids edges with highly negative curvature  and thus alleviates over-squashing, and enforces a maximum clique size of three, preventing uncontrolled densification that would accelerate over-smoothing.
TRIGON \cite{attali2025dynamic} further makes this principle task-adaptive by learning which candidate triangles (collected across the original graph, a $k$-NN feature similarity graph, and a Delaunay graph) should be retained through a differentiable selection module, yielding non-local triangulations that strengthen long-range connectivity.

Feature-similarity rewiring  (ComFy) \cite{rubio-madrigal2025gnns} rewires the graph by ranking candidate edge additions and deletions using a feature-affinity measure (typically cosine similarity) and applying the highest-ranked modifications. Community-aware variants constrain these modifications using an initial Louvain partition, so rewiring can strengthen within-community links or introduce controlled between-community connections without disrupting the original community layout.
In GOKU \cite{liang2025mitigating}, features intervene during sparsification: edge retention is biased by both feature similarity and a spectral importance signal (effective resistance–based), making structurally important, feature-consistent edges more likely to be kept, while feature-inconsistent ones are more likely to be removed.

JDR \cite{linkerhagner2025joint} targets regimes where both the topology and node features are noisy or misaligned: instead of relying on local similarity scores, it enforces global consistency between the graph and the feature geometry to obtain a topology more suitable for message passing. Concretely, it alternates between rewiring the graph and denoising the node features to maximize a spectral alignment between the top-$L$ eigenvectors of the adjacency matrix and the top-$L$ singular vectors of the original node features, thereby facilitating coherent long-range propagation.


\subsection{Practical Trade-offs: Structural vs. Feature-aware Rewiring, Efficiency, and Hyperparameter Sensitivity}
\label{sec:tradeoffs_signal_efficiency}

\paragraph{Performance attribution.}
A central practical issue is attribution: whether the observed gains truly reflect the effect of changing connectivity, or whether they partly arise from feature or model-dependent effects.
While feature-aware rewiring methods often deliver stronger empirical performance precisely because they exploit task-relevant signals encoded in the node attributes, their improvements can be harder to interpret causally.
Structural rewiring enables a more straightforward attribution, since the edited graph is determined from $G$ alone; this makes it easier to relate improvements to explicit connectivity patterns and to use rewiring as a controlled probe of the role of structure in message passing.
In contrast, feature-aware rewiring can be beneficial when the edges are not the most informative support for the task, but the interpretation of improvements becomes less specific: the rewiring signal is coupled to the feature space and may therefore act as an additional modeling step, so that improved accuracy cannot be attributed to topology reconstruction alone.

\paragraph{Hyperparameters.} The interpretability of structural-fix rewiring is further complicated by two recurring limitations: sensitivity to rewiring hyperparameters and computational overhead. Most structural-fix rewiring methods introduce additional design choices, most commonly an edit budget (the number of edges added and/or removed) or threshold rules \cite{UNDERSTANDING_bottlenecks,FOSR,BORF,resit,giraldo2023trade,barbero2024localityaware}. Empirically, the benefits of curvature-based rewiring on real-world benchmarks are highly sensitive to both training and rewiring hyperparameters and vary across graphs; reported SOTA gains often arise from favorable hyperparameter configurations rather than consistent improvements over the original topology \cite{tori2024effectiveness}. Only a subset of approaches aims to reduce this tuning burden, for instance, by calibrating curvature-based decisions through mixture modeling \cite{fesser2023mitigating} or by adopting parameter-free geometric constructions in feature space \cite{attali2024delaunay,attali2025dynamic}.

\paragraph{Complexity.} Finally, scalability is often constrained by the cost of the underlying descriptors: depending on the estimator, curvature-based criteria can scale quadratically to cubically in the number of edges \cite{UNDERSTANDING_bottlenecks,BORF}, effective resistance objectives can scale cubically in the number of nodes \cite{resit,liang2025mitigating}, and spectral criteria require Laplacian decompositions \cite{FOSR}. In contrast, several approaches rely on descriptors and constructions that are more efficient to compute and do not scale quadratically with the number of nodes \cite{DIGL,alon2020bottleneck,barbero2024localityaware,attali2024delaunay,linkerhagner2025joint}. As a result, these methods typically incur substantially lower overhead in practice and are easier to deploy on large graphs.

\begin{table*}[!h]
\centering
\caption{Comparison of various rewiring methods. OSQ: over-squashing; OSM: over-smoothing; UF: uses node features for rewiring; DT: rewiring performed during training (dynamic).}

\setlength{\tabcolsep}{6pt}
\renewcommand{\arraystretch}{1.1}
\begin{tabular}{llcccc l}
\toprule
\textbf{Models} & \textbf{Venues} & \textbf{OSQ} & \textbf{OSM} & \textbf{UF} & \textbf{DT} & \textbf{Measure} \\
\midrule
DIGL 
& [NeurIPS 2019] \cite{DIGL}
& \xmark & \xmark & \xmark & \xmark & PageRank \\

DropEdge 
& [ICLR 2020] \cite{rong2019dropedge}
& \xmark & \cmark & \xmark & \cmark & Random \\

PPRGo
& [KDD 2020] \cite{bojchevski2020scaling}
& \xmark & \xmark & \xmark & \xmark & PageRank \\

FA
& [ICLR 2021] \cite{alon2020bottleneck}
& \cmark & \xmark & \xmark & \xmark & Connectivity \\

SDRF 
& [ICLR 2022] \cite{UNDERSTANDING_bottlenecks}
& \cmark & \xmark & \xmark & \xmark & Balanced Forman curvature \\

RLEF 
& [IEEE AACCC 2022] \cite{banerjee2022oversquashing}
& \cmark & \xmark & \xmark & \xmark & Cheeger constant \\

DiffWire
& [LoG 2022] \cite{arnaiz2022diffwire}
& \cmark & \xmark & \cmark & \cmark & Commute Time and Spectral Gap \\

EGP 
& [LoG 2022] \cite{deac2022expander}
& \cmark & \xmark & \xmark & \xmark & Expander graph \\

FOSR 
& [ICLR 2023] \cite{FOSR}
& \cmark & \cmark & \xmark & \xmark & Spectral gap \\

BORF 
& [ICML 2023] \cite{BORF}
& \cmark & \cmark & \xmark & \xmark & Ollivier curvature \\

GTR 
& [ICML 2023] \cite{resit}
& \cmark & \xmark & \xmark & \xmark & Effective resistance \\

SJLR 
& [CIKM 2023] \cite{giraldo2023trade}
& \cmark & \cmark & \cmark & \cmark & Jost--Liu curvature \\

AFR-3 
& [LoG 2023] \cite{fesser2023mitigating}
& \cmark & \cmark & \xmark & \xmark & Augmented Forman curvature \\

PR-MPNNs 
& [ICLR 2024] \cite{qian2023probabilistically}
& \cmark & \xmark & \cmark & \cmark & Probabilistic learned rewiring \\

LASER 
& [ICLR 2024] \cite{barbero2024localityaware}
& \cmark & \xmark & \xmark & \xmark & Random walk \\

DR 
& [ICML 2024] \cite{attali2024delaunay}
& \cmark & \cmark & \cmark & \xmark & Delaunay triangulation \\

Pr Add/Delete
& [NeurIPS 2024] \cite{jamadandi2024spectral}
& \cmark & \cmark & \xmark & \xmark & Spectral gap \\

IPR-MPNN
& [NeurIPS 2024] \cite{qian2024probabilistic}
& \cmark & \xmark & \cmark & \cmark & Virtual nodes \\

CGP
& [LoG 2024] \cite{wilson2024cayley}
& \cmark & \xmark & \xmark & \xmark & Cayley expander \\

ComFy
& [ICLR 2025] \cite{rubio-madrigal2025gnns}
& \cmark & \cmark & \cmark & \xmark & Feature similarity \\

VN
& [ICLR 2025] \cite{southern2025understanding}
& \cmark & \xmark & \xmark & \xmark & Virtual node \\

JDR
& [ICLR 2025] \cite{linkerhagner2025joint}
& \cmark & \xmark & \cmark & \xmark & Spectral alignment \\

TRIGON
& [CIKM 2025] \cite{attali2025dynamic}
& \cmark & \cmark & \cmark & \cmark & Triangulation learning \\

GOKU
& [ICML 2025] \cite{liang2025mitigating}
& \cmark & \xmark & \cmark & \xmark & Effective resistance \\

FN
& [AAAI 2026] \cite{choi2025graph}
& \cmark & \xmark & \xmark & \xmark & Fractal node \\

\bottomrule
\end{tabular}
\label{Review}
\end{table*}

\subsection{Evaluation}
Rewiring should not be evaluated solely through downstream accuracy, since its primary goal is to modify the graph geometry to improve information propagation. In practice, one typically compares a fixed backbone GNN trained and evaluated on the original graph $G$ and on the rewired graph $G^{+}$ under a matched protocol. Beyond task-level scores, evaluation can include structure-level diagnostics that assess whether $G^{+}$ attenuates connectivity patterns known to exacerbate over-squashing and over-smoothing (e.g., shifts in edge-curvature distributions, decreases in effective resistance, or reductions in commute times), as well as model-level diagnostics that quantify whether long-range sensitivity has improved, for instance via a Jacobian-based sensitivity measure \cite{di2023over,saber2025over}. Task-level validation is typically reported on node and graph prediction benchmarks, with particular emphasis on heterophilic datasets \cite{lim2021large,platonov2023critical} and on long-range interaction benchmarks \cite{dwivedi2022long,miglior2026can,liang2026towards}, which directly stress the ability of GNNs to transmit signals across distant nodes.


\section{Discussion}

\paragraph{Recommendations.} We suggest articulating rewiring contributions through three complementary criteria.
First, the method should be motivated by the effective radius of the task, i.e., whether the predictive signal is expected to depend on information beyond a small neighborhood; when this is not the case, topology changes primarily targeting long-range communication may not be the most appropriate intervention.
Second, the evaluation should clarify the role of the input graph \cite{bechler-speicher2025position} by including a graph-agnostic baseline, such as DeepSets \cite{zaheer2017deep}, thereby distinguishing the benefits that truly arise from the relational structure from those attainable without the given connectivity. Additionally, a single rewiring method is unlikely to be appropriate across diverse applications, graph families, and tasks. By altering aggregation neighborhoods, rewiring implicitly determines which relations are treated as informative for the downstream objective. This choice depends on the application-specific meaning of edges and on the task’s characteristic dependency scale in the graph. Accordingly, a rule that benefits one regime may harm another by violating domain constraints or by aggregating incompatible feature evidence.
Third, reported gains should be supported by a joint analysis of effectiveness, computational cost, and attribution: downstream improvements should be reported together with the edit rewiring budget, runtime and memory overhead, and matched-budget controls (e.g., random or degree-preserving rewiring baselines, and ablations for feature-driven constructions) so that the source of the improvement is identifiable and comparisons across methods are meaningful.



 \paragraph{Future directions.} Building on the recommendations above, we outline broader research directions in which rewiring serves not only as an optimization tool, but also as a controlled structural intervention for understanding how topology shapes information propagation and learning in GNNs. A key direction is to better understand the interplay between graph structure and node features. Since message passing couples features through the graph, connectivity changes directly modulate how attribute information is propagated and combined, which can either reinforce or conflict with the task signal. An important open problem is to clarify the respective roles of structure and features in long-range learning: when feature side mechanisms already convey the non-local signal \cite{sketched-random-features}, when rewiring is indispensable, and how to combine both under matched budgets and constraints. More broadly, rewiring enables controlled neighborhood perturbations to study this interplay; yet principled criteria for predicting when structure and features are aligned or when their mismatch induces harmful mixing remain largely open \cite{gomes2022graph}.

Additionally, graph structure can also shape generalization through the stability of message passing, by controlling how strongly neighborhood aggregation propagates and amplifies perturbations across layers. Recent analyses make this link explicit by expressing bound constants in terms of structural quantities such as degree heterogeneity and spectral properties of the graph filter \cite{ju2023generalization}. This suggests a promising direction: leverage such bound-relevant structural proxies to anticipate when topology is likely to degrade generalization, and to guide rewiring toward controlling propagation amplification rather than optimizing accuracy alone.

\bibliographystyle{named}
\bibliography{ijcai26}

\end{document}